# A Neuro-Fuzzy Technique for Implementing the Half-Adder Circuit Using the CANFIS Model

Sachin Lakra[1], T. V. Prasad[2], Deepak Sharma[3], Shree Harsh Atrey[4], Anubhav Sharma[5]

[1]Department of Information Technology, Manav Rachna Rachna College of Engineering,
Faridabad, Haryana, India
[2]Department of Computer Science and Engineering, Lingaya's Institute of Management and Technology,
Faridabad, Haryana, India
[3]Department of Information Technology, Manav Rachna Rachna College of Engineering,
Faridabad, Haryana, India
[4]Department of Computer Science and Engineering, Manav Rachna Rachna College of Engineering,
Faridabad, Haryana, India
[5]Department of Information Technology, Lingaya's Institute of Management and Technology,
Faridabad, Haryana, India

**Abstract -** *A Neural Network, in general, is not considered to be a good solver of mathematical and binary arithmetic problems. However, networks have been developed for such problems as the XOR circuit. This paper presents a technique for the implementation of the Half-adder circuit using the CoActive Neuro-Fuzzy Inference System (CANFIS) Model and attempts to solve the problem using the NeuroSolutions 5 Simulator. The paper gives the experimental results along with the interpretations and possible applications of the technique.*

## 1. Introduction

The authors performed experiments using Simple Modular Neural Networks and CANFIS Model based Neuro-Fuzzy Networks. The binary arithmetic problem [8] of a Half-adder circuit was attempted to be solved using each of these two types of Neural Networks. The first approach was based on Simple Modular Neural Networks and attempted to solve the problem using the Java Object Oriented Neural Engine (Joone) Simulator. The results obtained in this approach were found to be unsuitable for solving the Half-adder circuit problem.

The second approach was based on the CANFIS model and attempted to solve the problem [7] using the NeuroSolutions 5 Simulator. This approach uses fuzzy preprocessing of the training data followed by use of Modular Neural Networks to solve the given problem giving a much better performance in terms of the number of epochs required to achieve a given Mean Squared Error (MSE). The results of the experiments using the second approach found that the CANFIS model was suitable for solving the Half-adder circuit problem.

Section 1 introduces the paper. Section 2 defines the problem being attempted in the paper. Section 3 describes the Methods used in the experiments, along with the data used and the conditions of the experiments. Section 4 presents the results of the experiments, in the form of Training, Performance and Testing Reports. Section 5 discusses and interprets the results giving a relationship between the parameters involved. Section 6 concludes the paper.

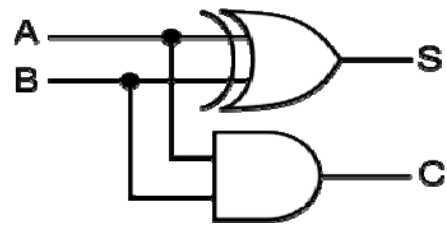

**Figure 1:The Half-Adder circuit
(Source: http://202.44.12.92/bellab)**

## 2. Problem Statement

The problem solved in this paper may be stated as the implementation of the binary Half-Adder circuit using a neuro-fuzzy approach based on the (CANFIS) Model using the NeuroSolutions neural network simulator. The Half-adder circuit used is shown in Figure 1.

## 3. Methods

### 3.1 Unsuitability of the Simple Modular Neural Network Approach

The first approach tested for solving the problem involved the use of a Modular Neural Network to create a binary Half-Adder circuit using the Java Object Oriented Neural Engine (Joone) Neural Network simulator. The neural network consisted of two Nested ANN's. The first Nested ANN was a XOR gate trained to a final RMSE of approximately 0.0009. The second Nested ANN was an AND gate, again trained to a final RMSE of approximately 0.0009. The Modular ANN consisted of a network similar to the Half-adder circuit of Figure 1 and is shown in Figure 2.

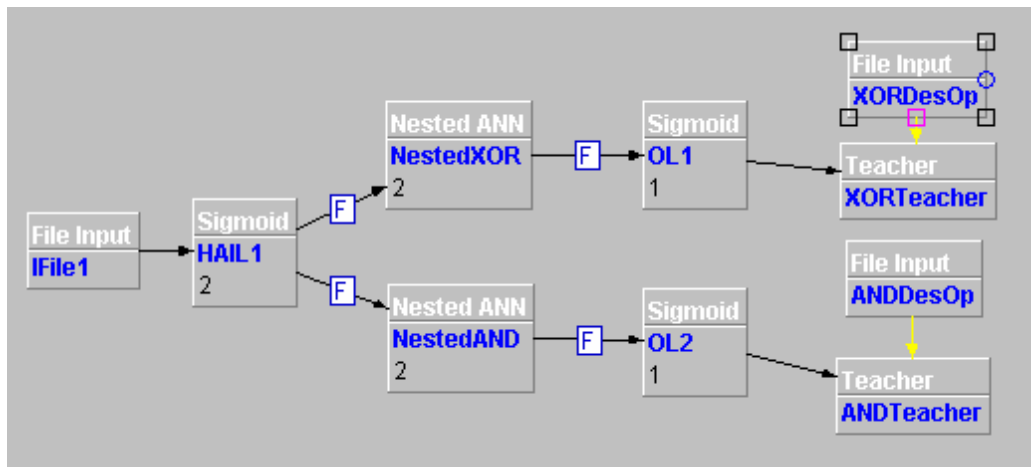

**Figure 2: The Half-adder circuit implemented in Joone.**

The approach was found to be unsuitable as the RMSE of the Half-adder Modular ANN did not fall below approximately 0.35 and the training of the neural network was unsuccessful. The test results obtained were far from the sum and carry outputs of the Half-adder circuit. This approach was therefore abandoned.

### 3.2 The CANFIS Approach

The CANFIS model integrates adaptable fuzzy inputs with a modular neural network to rapidly and accurately approximate complex functions. Fuzzy inference systems are also valuable as they combine the explanatory nature of rules (membership functions) with the power of "black box" neural networks [2-5, 9].

The technique of cross validation has been used in the training of the network. Cross validation is a highly recommended method for stopping network training. This method monitors the error on an independent set of data and stops training when this error begins to increase. This is considered to be the point of best generalization.

The testing set is used to test the performance of the network. Once the network is trained the weights are then frozen, the testing set is fed into the network and the network output is compared with the desired output.

### 3.3 Data used

The following is the data used in the Experiments.

**Training data:** The training data used is the Truth Table for the Half-Adder circuit and is shown in Table 1. X and Y are the two inputs given to the CANFIS network, and S and C are the Sum and Carry outputs as desired to be obtained from the network.

**Table 1: Training Data**

| X | Y | S | C |
|---|---|---|---|
| 0 | 0 | *0* | *0* |
| 0 | 1 | *1* | *0* |
| 1 | 0 | *1* | *0* |
| 1 | 1 | *0* | *1* |

(Source: http://www.cs.umd.edu)

**Cross validation data:** The data used for Cross Validation consists of the same two inputs and the same two outputs as in the training data. However, X and Y are given as approximations to the values 1 and 0 to introduce variation in the only four discrete combinations possible otherwise. The Cross Validation data is shown in Table 2.

**Testing data:** The testing data is the data on which the performance of the Half-adder CANFIS network is tested and is shown in Table 3.

**Table 2: Cross Validation Data**

| X | Y | S | C |
|---|---|---|---|
| 0.05 | 0.03 | *0* | *0* |
| 0.09 | 0.98 | *1* | *0* |
| 0.06 | 1.06 | *1* | *0* |
| 1.02 | 0.96 | *0* | *1* |
| 0.97 | 0.035 | *1* | *0* |
| 0.99 | 0.97 | *0* | *1* |
| 0.055 | 0.98 | *1* | *0* |
| 1.01 | 0.03 | *1* | *0* |
| 1.04 | 0.99 | *0* | *1* |

The data in Tables 1, 2 and 3 is the same for each of Experiments 1, 2 and 3.

**Table 3: Testing Data**

| X | Y | S | C |
|---|---|---|---|
| 0.07 | 0.02 | *0* | *0* |
| 0.09 | 0.99 | *1* | *0* |
| 1.045 | 0.03 | *1* | *0* |
| 0.08 | 0.01 | *0* | *0* |
| 0.98 | 0.02 | *1* | *0* |
| 0.975 | 0.98 | *0* | *1* |

### 3.4 Experimental conditions

The experimental conditions are shown in Table 4 and are the same for each of Experiments 1, 2 and 3, except for the number $n_{mf}$ of membership functions per input. The network created using the NeuroSolutions 5 Simulator is shown in Figure 3.

**Table 4: Experimental conditions**

| Parameter | Value | |
|---|---|---|
| **Half-adder CANFIS Network** | | |
| Membership Function | Bell | |
| Fuzzy Model | TSK | |
| No. of Input Processing elements | 2 | |
| No. of Exemplars | 4 | |
| | | |
| **Output Layer** | | |
| No. of Output Processing elements | 2 | |
| Transfer function | SigmoidAxon | |
| Learning Rule | Momentum with Step size=1 and momentum = 0.6 | |
| | | |
| **Type of Learning** | Supervised | |
| **Number of Epochs** | 1000 | |
| **Termination conditions** | based on Mean Square Error using Cross Validation data to terminate on an increase in MSE. | |
| **Conditions during Training** | Cross Validation to be used | |
| | Initial weights to be randomized. | |
| **Conditions during Testing** | Load Best weights | |
| | Use regression. | |
| **Number of Membership Functions per input n (only parameter that differs for each Experiment)** | Experiment 1 | n = 2 |
| | Experiment 2 | n = 3 |
| | Experiment 3 | n = 4 |

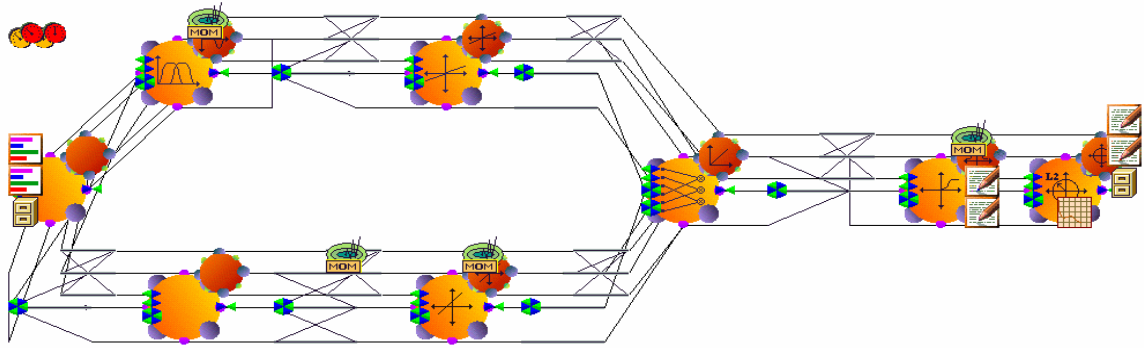

**Figure 3: The Half-adder CANFIS Model based Neuro-Fuzzy Network developed using NeuroSolutions 5**

**Table 5: Training Report**

| Best Network | Experiment 1 | | Experiment 2 | | Experiment 3 | |
|---|---|---|---|---|---|---|
| Epoch # | Training | Cross Validation | Training | Cross Validation | Training | Cross Validation |
| 1 | 0.1083899 | 0.1020836 | 0.1045081 | 0.0985990 | 0.0926004 | 0.0892539 |
| 200 | 0.0319968 | 0.0344914 | 0.0129120 | 0.0126704 | 0.0146223 | 0.0138344 |
| 400 | 0.0045845 | 0.0052281 | 0.0053343 | 0.0055623 | 0.0046065 | 0.0043432 |
| 600 | 0.0010367 | 0.0012859 | 0.0027516 | 0.0029660 | 0.0021402 | 0.0020216 |
| 800 | 0.0003791 | 0.0005171 | 0.0015122 | 0.0016577 | 0.0012175 | 0.0011558 |
| 1000 | 0.0001761 | 0.0002683 | 0.0009010 | 0.0010028 | 0.0007791 | 0.0007449 |
| Min MSE (1000 ep) | 0.0001761 | 0.0002683 | 0.0009010 | 0.0010028 | 0.0007791 | 0.0007449 |
| Final MSE (1000 ep) | 0.0001761 | 0.0002683 | 0.0009010 | 0.0010028 | 0.0007791 | 0.0007449 |

## 4. Results

### 4.1 Training Reports

The Training Reports are shown in Graphs 1, 2 and 3 and Table 5.

### 4.2 Performance and Testing Reports

The Performance Report is shown in Table 6 and the Testing reports are shown in Graphs 4, 5 and 6 and Table 7.

## 5. Discussion and Interpretation

The implementation of the Half-adder circuit is possible using a neural network [1]. However, the problem has not yet been attempted to be solved using a neuro-fuzzy approach, and, in particular, using the CANFIS Model.

From Graphs 1, 2 and 3, it can be interpreted that initial training upto 200 epochs is poor with the number $n_{mf}$ of Membership Functions per input being 2 as in Experiment 1, and improves with an increase in $n_{mf}$ from 3 to 4, as in Experiments 2 and 3, respectively. However, from Table 5, the Minimum MSE at the end of 1,000 epochs is least for $n_{mf}=2$ in Experiment 1 and highest for $n_{mf}=3$ in Experiment 2, but improves a bit for $n_{mf}=4$ (as in Experiment 3) as compared to Experiment 2. In further experiments, it was found that the minimum MSE becomes many times higher than that of Experiment 1. Also, a low value of 0.001 of the MSE is achieved in Experiment 1 at 593 epochs, in Experiment 2 at 922 epochs and in Experiment 3 at 844 epochs. Further experiments found that an MSE of 0.001 was obtained at 1464 epochs for $n_{mf}=5$ and at 1169 epochs for $n_{mf}=6$. From these, it can be concluded that the best overall training is given by Experiment 1.

Thus, for the purpose of training, it can be concluded that

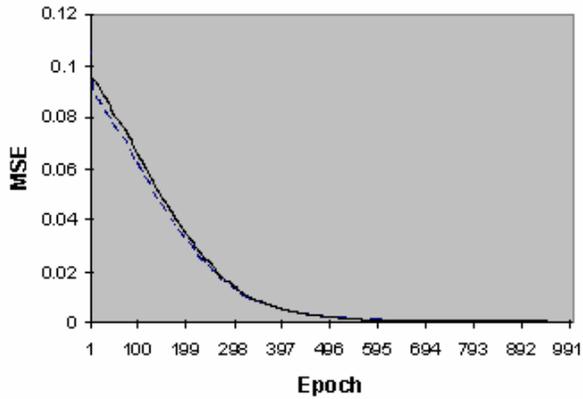

**Graph 1: Experiment 1 Training Report**

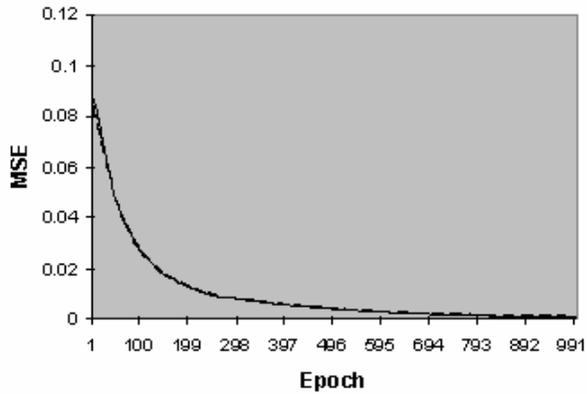

**Graph 2: Experiment 2 Training Report**

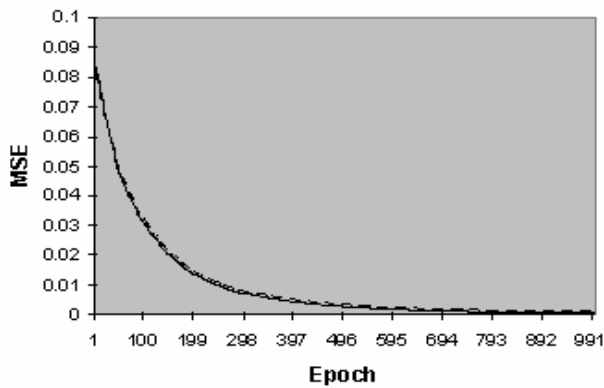

**Graph 3: Experiment 3 Training Report**

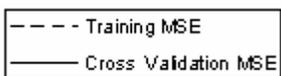

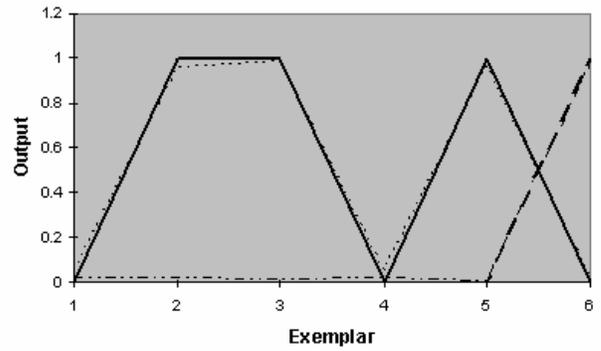

**Graph 4: Experiment 1 Testing Report**

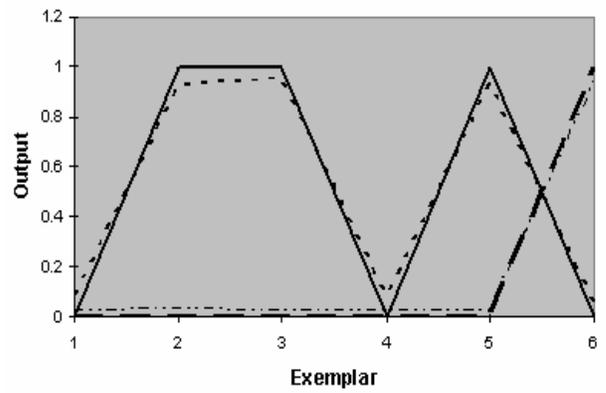

**Graph 5: Experiment 2 Testing Report**

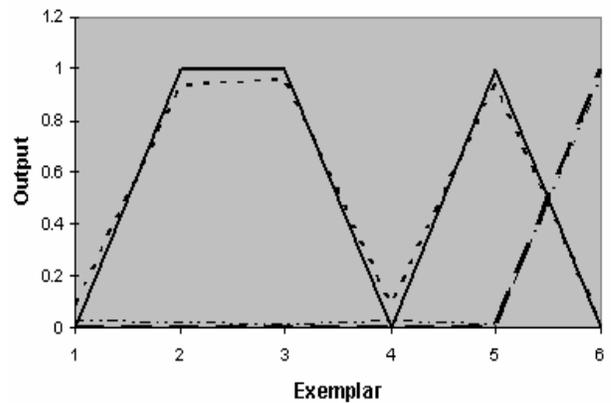

**Graph 6: Experiment 3 Testing Report**

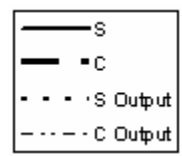

## Table 6: Performance Report of the Network

| Performance | Experiment 1 | | Experiment 2 | | Experiment 3 | |
|---|---|---|---|---|---|---|
| | S | C | S | C | S | C |
| **MSE** | 0.0014003 | 0.0003647 | 0.0048026 | 0.0010965 | 0.0043399 | 0.0007029 |
| **NMSE** | 0.0056014 | 0.0026258 | 0.0192107 | 0.0078950 | 0.0173599 | 0.0050609 |
| **MAE** | 0.0351559 | 0.0175923 | 0.0676219 | 0.0305150 | 0.0614169 | 0.0238830 |
| **Min Abs Error** | 0.0175748 | 0.0079792 | 0.0477134 | 0.0227878 | 0.0211649 | 0.0153816 |
| **Max Abs Error** | 0.0493182 | 0.0311184 | 0.0850152 | 0.0589271 | 0.0877884 | 0.0485532 |
| **R** | 0.9996588 | 0.9999263 | 0.9994543 | 0.9999832 | 0.9985683 | 0.9999553 |

Legend: MSE = Mean Squared Error;   NMSE = Normalized Mean Squared Error
MAE = Mean Absolute Error;   r = Linear Correlation Coefficient

## Table 7: Testing Report

| Inputs | | Desired | | Experiment 1 | | Experiment 2 | | Experiment 3 | |
|---|---|---|---|---|---|---|---|---|---|
| X | Y | S | C | S Output | C Output | S Output | C Output | S Output | C Output |
| 0.07 | 0.02 | 0 | 0 | 0.048782 | 0.017349 | 0.084328 | 0.024663 | 0.087297 | 0.023342 |
| 0.09 | 0.99 | 1 | 0 | 0.957826 | 0.021271 | 0.925878 | 0.028877 | 0.930268 | 0.016920 |
| 1.045 | 0.03 | 1 | 0 | 0.982425 | 0.011152 | 0.952286 | 0.023126 | 0.957306 | 0.015381 |
| 0.08 | 0.01 | 0 | 0 | 0.049318 | 0.016682 | 0.085015 | 0.024707 | 0.087788 | 0.023284 |
| 0.98 | 0.02 | 1 | 0 | 0.966619 | 0.007979 | 0.934195 | 0.022787 | 0.940173 | 0.015815 |
| 0.975 | 0.98 | 0 | 1 | 0.019705 | 0.968881 | 0.048747 | 0.941072 | 0.021164 | 0.951446 |

$$Min(MSE) \propto \frac{1}{n_{mf} \times n_e}, n_{mf} > 0, n_e > 0 \quad (1)$$

where,   MSE = Mean Squared Error,
$n_{mf}$ = number of membership functions per input,
$n_e$ = number of epochs in which the minimum MSE is achieved.

Besides this, as can be seen from Table 7, the outputs obtained from the experiments are very close to 0's and 1's but are not exactly 0's and 1's inspite of the training data being exactly the same as the truth table for the Half-adder circuit. This does not alter the Half-adder circuit's truth table in any way as the voltages forming a square wave in a computer's circuit are not precisely 1's and 0's but are values *near* 1's and 0's.

## 6. Conclusion

Thus, a Half-adder circuit can be developed using the CANFIS Model. The testing data used and the results obtained from the experiments described in this paper point towards the creation of electronic circuits using neuro-fuzzy networks and can open a vista of new possibilities in the area of the development of neuro-fuzzy computers[6] that can be used as an intelligent replacement of current computers.

Another application of such electronic circuits is in the area of digitization of electrical analog signals by neuro fuzzy instruments based on fuzzy sampling of the analog signal.